\documentclass[letterpaper,10pt,conference]{ieeeconf}
\IEEEoverridecommandlockouts
\usepackage{amsmath,amssymb,amsfonts}
\usepackage{graphicx}
\usepackage{array}
\usepackage{booktabs}
\usepackage[table]{xcolor}
\usepackage{textcomp}
\usepackage{url}
\usepackage{cite}
\usepackage{xspace}
\providecommand{\etal}{\emph{et al.}\xspace}

\newif\ifanon
\anonfalse    

\title{RAPA: Ray-Aware Physics Augmentation\\
for Robust Camera–LiDAR Calibration}

\ifanon
  \author{\mbox{}}
\else
  \author{Kyung-Rin Oh$^{*}$, Ji-Hoon Kim$^{*}$, TaeJoon Ha, and Jae-Sang Hyun$^{\dagger}$%
    \thanks{$^{*}$These authors contributed equally to this work.}%
    \thanks{This work has been submitted to the IEEE for possible publication.
    Copyright may be transferred without notice, after which this version
    may no longer be accessible.}%
    \thanks{This research was supported by the Technology Innovation Program,
funded by the Ministry of Trade, Industry and Resources (MOTIR, Korea),
under Project RS-2025-25447257, ``Development of AI autonomous continuous
production system technology for gas turbine blade maintenance and
regeneration for power generation'' (contribution rate: 35\%).
This research was also supported by National Research Foundation of
Korea (NRF) grants funded by the Korean government (MSIT) under
Projects RS-2026-25505230 (contribution rate: 30\%) and
RS-2026-25498577 (contribution rate: 30\%), and by the Yonsei Institute
for Embodied Intelligence, Yonsei University (contribution rate: 5\%).}
    \thanks{$^{\dagger}$Corresponding author: {\tt\small hyun.jaesang@yonsei.ac.kr}}%
    \thanks{All authors are with the Department of Mechanical Engineering,
Yonsei University, Seoul, Republic of Korea. Jae-Sang Hyun is also with
the Yonsei Institute for Embodied Intelligence, Yonsei University,
Seoul, Republic of Korea.}%
  }
\fi

\begin{document}
\maketitle
\thispagestyle{empty}
\pagestyle{empty}

\begin{abstract}
Camera--LiDAR calibration is essential for reliable perception in robotic and autonomous systems. Most target-based calibration methods rely on checkerboard observations, but accurate checkerboard feature extraction from LiDAR point clouds remains challenging due to measurement sparsity. Existing approaches either require manual annotation of 3D corners or depend on geometric feature extraction methods, both of which can become unreliable when checkerboard features are insufficiently observed. In this paper, we formulate checkerboard localization as a 2D detection problem using cylindrical LiDAR reflectivity images. To overcome the limited availability of labeled training data, we propose ray-aware physics augmentation (RAPA), which generates diverse synthetic samples from a single checkerboard observation. The generated data are used to improve detector robustness and generalization under varying sensing conditions. Across four independently captured indoor and outdoor locations, the proposed synthetic augmentation exceeds the strongest evaluated image-space augmentation baseline by 16.1--43.1 percentage points in mAP@0.5. Compared with training-free target-based methods on 242 observations, RAPA yields the largest number of valid target localizations, with 96.4\% of its returned detections yielding valid target poses. These reliable localizations support downstream camera--LiDAR calibration across all four evaluated locations. Overall, RAPA addresses a key bottleneck in target-based calibration by enabling automated LiDAR-side target localization with only a single annotated target observation.
\end{abstract}


\section{Introduction}

\begin{figure*}[!t]
    \centering
     \includegraphics[
        width=\textwidth,
        trim=0cm 0.4cm 1cm 0cm,
        clip
    ]{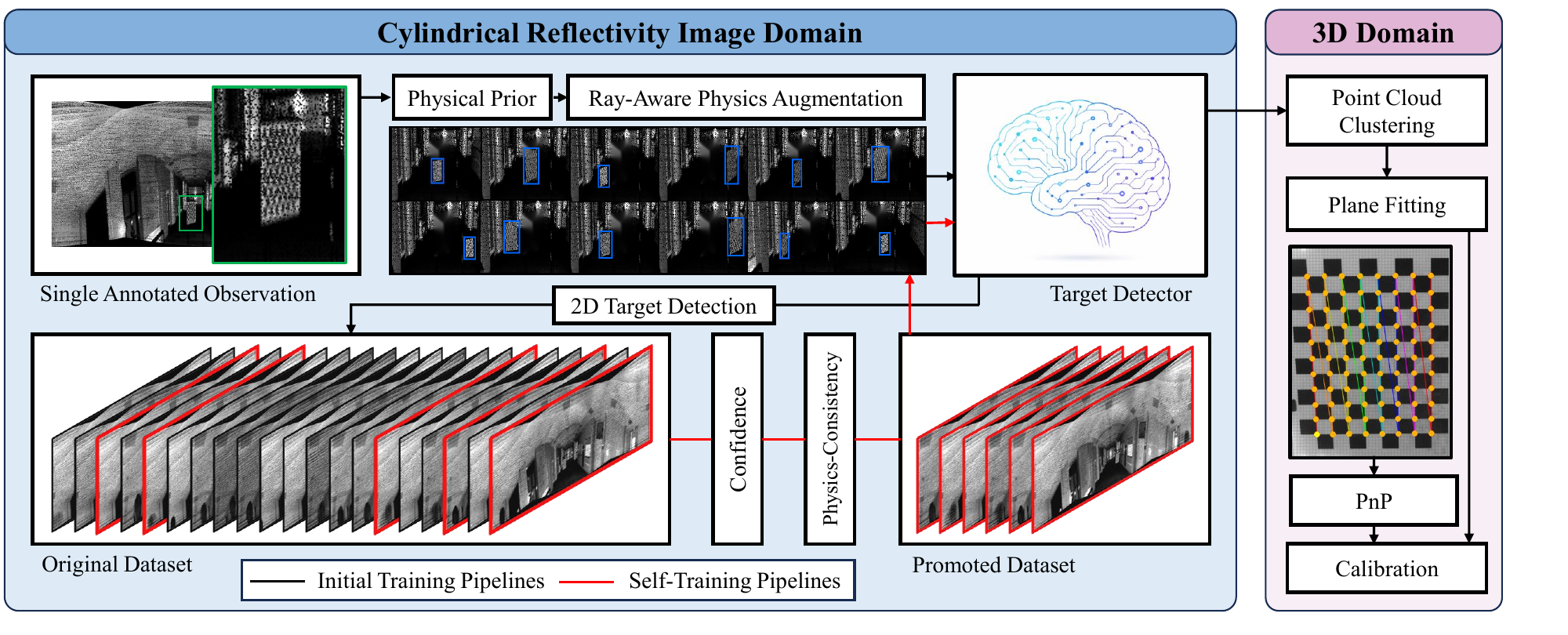}
    \caption{RAPA overview. Geometry and reflectivity priors from one
    annotated observation and a sensor--environment range-response
    prior generate ray-consistent synthetic targets. One self-training
    round refines the detector on unlabeled observations. Detected
    regions guide 3D board-pose recovery for camera--LiDAR calibration.}
    \label{fig:framework}
\end{figure*}

Cameras and LiDARs provide complementary appearance and metric geometry
for autonomous vehicles, mobile robots, and mapping
systems~\cite{huang2022multimodal,Geiger2013IJRR,Sun2020Waymo}.
Extrinsic calibration aligns these measurements for perception and
localization~\cite{pandey2012automatic,Geiger2012}.
Checkerboards provide repeatable geometric
constraints~\cite{Zhang2004extrinsic,Geiger2012} and are reliably detected
in camera images~\cite{Zhang2000}. In LiDAR, sparse and non-uniform
sampling can leave their boundaries and internal pattern only partially
observed, particularly at long ranges or unfavorable
viewpoints~\cite{Wang2017reflectance,Fu2023diffcal,cui2020acsc}.
This makes LiDAR-side target localization a bottleneck in calibration
automation.

We formulate checkerboard localization as 2D detection in cylindrical
reflectivity images (Fig.~\ref{fig:framework}). This representation
preserves the angular scan structure and black--white target contrast;
the detected region identifies 3D returns for board-pose recovery.

Learning a robust detector in this domain is challenging when
annotated calibration-target data are scarce. Acquiring observations
over diverse ranges, viewpoints, and backgrounds requires dedicated
data collection, while manual target annotation is
labor-intensive~\cite{Wang2023TLSAlignment}.
This creates a practical tension: target localization should reduce
manual intervention in calibration, but training the detector should
not require a large collection of manually localized targets.
To train with limited annotation, ray-aware physics augmentation
(RAPA) combines geometry and reflectivity priors from one annotated
observation with a range-response prior estimated from separate static
surfaces. It renders virtual target poses using per-ray intersection
range and incidence angle. One round of self-training adds reliable
unlabeled observations without further manual
annotation~\cite{Lee2013Pseudolabel}.

The resulting detector automates target localization in subsequent
observations and supplies regions for geometric pose recovery.
We evaluate this contribution to camera--LiDAR calibration across four
indoor/outdoor locations, with additional targets and sensors testing
the framework's applicability.

Our main contributions are listed as follows:
\begin{itemize}
    \item We formulate LiDAR checkerboard localization as 2D detection in cylindrical reflectivity images, providing a target-specific region for subsequent 3D geometric fitting under sparse point sampling.

    \item We propose ray-aware physics augmentation combining
    target-specific priors from one annotated observation with an
    empirical sensor--environment range-response prior, ray-wise
    rendering, and one round of self-training.

    \item We advance camera--LiDAR calibration automation by linking
    automatic LiDAR target localization to 3D pose recovery after one
    annotated observation. Verified target correspondences support
    calibration across all four evaluated locations.
\end{itemize}

\section{Related Work}

\subsection{Camera--LiDAR Calibration}

Target-free calibration uses cross-modal
appearance~\cite{huang2024mias,Jung2026TLCCalib},
reflectivity~\cite{pandey2012automatic}, edges~\cite{liao2023secalib}, or
geometric structures~\cite{Iyer2018CalibNet,huang2026dstcalib}, with
method-specific assumptions about initialization and cross-modal
feature matching.
Target-based methods use known geometry: Zhang and
Pless~\cite{Zhang2004extrinsic} related checkerboard planes to laser
measurements, Geiger \etal~\cite{Geiger2012} used multiple checkerboards
for single-shot calibration, and Zhou \etal~\cite{Zhou2018automatic}
extracted 3D boundary and plane correspondences. Reflectance and
temporal cues help verify candidates~\cite{cui2020acsc}, but sparse
returns can still limit geometric support~\cite{Hu2022PDV}.
These approaches provide repeatable geometric constraints once the
target is localized, but locating that target remains a distinct
requirement. Manual region selection can provide the required support,
while automatic methods must distinguish the board from competing
planar structures before extracting correspondences. RAPA supplies a
learned target region while retaining geometric fitting for 3D pose
estimation.

\subsection{LiDAR Intensity and Reflectivity}
Wang \etal~\cite{Wang2017reflectance} used checkerboard intensity
contrast to fit a board model and recover 3D corners; Fu
\etal~\cite{Fu2023diffcal} jointly used LiDAR intensity and camera
observations through a differentiable checkerboard model. We use
this contrast to synthesize detector-training observations.
Intensity also depends on range, incidence angle, and sensor response.
Viswanath \etal~\cite{Viswanath2025reflectivity} compensated for these
dependencies for semantic segmentation. Such compensation seeks a
consistent representation of a surface despite changes in sensing
geometry. Our objective is to predict how a target's measured appearance
changes at a new pose, so RAPA incorporates these dependencies into
forward rendering. Its empirical sensor--environment range-response
prior captures the response under the acquisition conditions without
assuming a universal material-reflectance law.

\subsection{Learning with Limited Calibration Data}

Learning a calibration-target detector in LiDAR is challenging because annotated
LiDAR observations are typically scarce~\cite{Xiao2023}.  Calibration
datasets usually contain only a limited number of annotated target
observations. As a result, the limited training diversity makes it
difficult to capture appearance variations across different sensing
conditions. Consequently, learning detectors that generalize to unseen
observations remains challenging~\cite{Wang2017reflectance,Fu2023diffcal}.

Cylindrical reflectivity images provide an alternative representation
that preserves the LiDAR scan structure and exposes the black--white
target contrast. Conventional image augmentations~\cite{Hu2019DeepCharuco,Shorten2019} perform pixel-level transformations without preserving the underlying ray geometry and angular sampling pattern of cylindrical LiDAR scans. As a result, the generated target appearances are geometrically inconsistent with the LiDAR sensing process. They also fail to model appearance variations caused by range, incidence angle, and sampling sparsity~\cite{Fan2021RangeDet}. In contrast, our augmentation preserves the underlying ray geometry while generating synthetic target observations.

\section{Proposed Method}

\begin{figure*}[t]
    \centering
    \includegraphics[width=\textwidth]
    {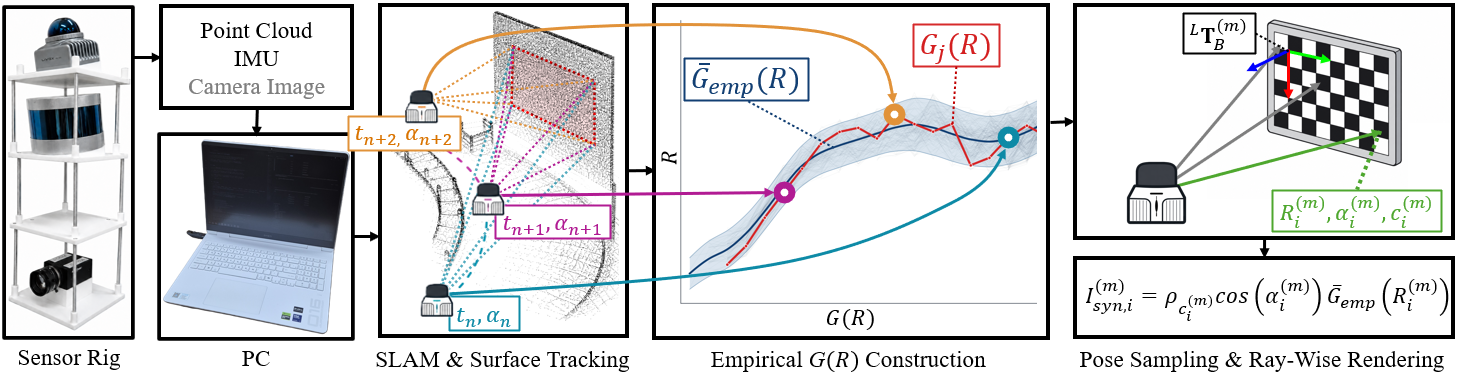}
    \caption{Automated construction and ray-wise application of the
    empirical range-response prior $G(R)$. Synchronized LiDAR--IMU
    scans are registered, and persistent planar surfaces are tracked
    across scans. Their incidence-corrected measurements are used to
    obtain retained per-track response curves $G_j(R)$. Their
    distancewise arithmetic mean is interpolated and mildly smoothed
    by a log-domain cubic P-spline to obtain the fixed response
    $\bar{G}_{\mathrm{emp}}(R)$. The shaded band denotes the sample
    standard deviation across retained tracks. For synthetic
    observation $m$, the smoothed response is evaluated
    at each ray--board intersection range $R_i^{(m)}$ and combined with
    the corresponding incidence angle $\alpha_i^{(m)}$ and checker
    class $c_i^{(m)}$. The LiDAR sensor and environment-group indices
    are omitted for clarity.}
    \label{fig:gr_estimation}
\end{figure*}

The single annotated target observation, or \emph{good frame}, supplies
the board geometry and reflectivity priors. RAPA follows
Fig.~\ref{fig:framework} to render training samples, refine the detector
through self-training, and recover target poses from detections.
Additional inputs are known target dimensions, a pretrained detector,
an unlabeled pool, and separate LiDAR--IMU recordings for the
range-response prior (Fig.~\ref{fig:gr_estimation}).

\subsection{Cylindrical Reflectivity Representation}

A LiDAR return is represented by
\(\mathbf{p}_i=(x_i,y_i,z_i,I_i)\), where \(I_i\) denotes the
intensity reported by the sensor. Its range, azimuth, and elevation
are computed as
\begin{equation}
    \begin{aligned}
    R_i &= \sqrt{x_i^2+y_i^2+z_i^2}, \\
    \theta_i &= \operatorname{atan2}(y_i,x_i), \\
    \phi_i &= \sin^{-1}\left(\frac{z_i}{R_i}\right).
    \end{aligned}
\end{equation}

Given fixed angular bounds
\([\theta_{\min},\theta_{\max}]\) and
\([\phi_{\min},\phi_{\max}]\), the return is projected onto a
cylindrical image of width \(W_I\) and height \(H_I\) as
\begin{equation}
    \begin{aligned}
    u_i &=
    \left\lfloor
    \frac{\theta_i-\theta_{\min}}
    {\theta_{\max}-\theta_{\min}}(W_I-1)
    \right\rfloor, \\
    v_i &=
    \left\lfloor
    \frac{\phi_{\max}-\phi_i}
    {\phi_{\max}-\phi_{\min}}(H_I-1)
    \right\rfloor.
    \end{aligned}
\end{equation}

Each valid pixel stores the intensity \(I\), range \(R\), validity
mask \(M\), and corresponding 3D coordinates \((x,y,z)\). The
intensity channel is converted into a grayscale reflectivity image
for detector training, while the range, validity, and 3D coordinate
channels retain the pixel-to-ray correspondence required for
ray-aware rendering and downstream board-pose estimation. 

\subsection{Ray-Aware Physics Augmentation from a Single Annotated Observation}
\subsubsection{Board geometry and checker alignment}
For the \emph{good frame}, the operator specifies the target using
corner points or a target-region seed in a single observation. The
selected 3D returns are used to fit the board plane automatically;
known target dimensions determine its center, normal, in-plane axes,
and physical support. Projecting this support provides the detector
bounding box. The resulting board pose \({}^{L}\mathbf{T}_{B}\) maps
the board coordinate frame \(B\) to the LiDAR coordinate frame \(L\):
\begin{equation}
    {}^{L}\mathbf{T}_{B}=
    \begin{bmatrix}
        \mathbf{a}_u & \mathbf{a}_v & \mathbf{n} & \mathbf{c}\\
        0 & 0 & 0 & 1
    \end{bmatrix},
\end{equation}

where \(\mathbf{c}\) is the board center,
\(\mathbf{a}_u\) and \(\mathbf{a}_v\) are its in-plane axes, and
\(\mathbf{n}\) is the board normal. The board has known dimensions $W_B$ and $H_B$, denoting the board width and height, with square size $d$.

For a unit ray direction \(\mathbf{d}\), the intersection with the board plane is
\begin{equation}
    \lambda=
    \frac{\mathbf{n}^{\top}\mathbf{c}}
    {\mathbf{n}^{\top}\mathbf{d}},
    \qquad
    \mathbf{p}=\lambda\mathbf{d}.
\end{equation}

The board-local coordinates of the intersection are
\begin{equation}
    u=(\mathbf{p}-\mathbf{c})^{\top}\mathbf{a}_u,
    \qquad
    v=(\mathbf{p}-\mathbf{c})^{\top}\mathbf{a}_v.
\end{equation}

A ray belongs to the physical board support when
\begin{equation}
    \lambda > 0, \quad |u| \leq W_B/2, \quad |v| \leq H_B/2.
\end{equation}

Since the board pose does not determine the printed checker
alignment, we additionally estimate the in-plane offsets
$(\delta_u,\delta_v)$ and binary phase $q$. For each candidate, the checker indices are
\begin{equation}
    \resizebox{0.9\columnwidth}{!}{$
    n_u=
    \left\lfloor
    \frac{u+W_B/2+\delta_u}{d}
    \right\rfloor,
    \qquad
    n_v=
    \left\lfloor
    \frac{H_B/2-v+\delta_v}{d}
    \right\rfloor.
    $}
\end{equation}

The white-square mask is defined by
\begin{equation}
\mathcal{M}_{\mathrm{white}}
=
\left[(n_u+n_v+q)\bmod2=0\right].
\end{equation}

The in-plane offsets and binary phase are selected by grid search as
\begin{equation}
\resizebox{0.9\columnwidth}{!}{$
(\delta_u^{*},\delta_v^{*},q^{*})
=
\operatorname*{arg\,max}_{\delta_u,\delta_v,q}
\left[
\operatorname{median}(I_{\mathrm{white}})
-
\operatorname{median}(I_{\mathrm{black}})
\right].
$}
\end{equation}

\subsubsection{Sensor range-response prior}
A standard LiDAR radiometric model expresses the received intensity as
\begin{equation}
    I(R,\alpha,\rho)
    \propto
    \eta(R)\frac{\rho\cos(\alpha)}{R^2},
\end{equation}
where $R^{-2}$ represents ideal geometric attenuation and
$\eta(R)$ captures additional range-dependent optical and sensor
effects~\cite{jelalian1980laser,Viswanath2025reflectivity}.
For a commercial LiDAR, however, the reported intensity is already
affected by proprietary sensor processing, making these range-dependent
components difficult to identify separately. We therefore model their
combined effect empirically as a range-response function $G(R)$.
Specifically, $G(R)$ is estimated from tracked static surfaces, and
the resulting arithmetic-mean response
$\bar{G}_{\mathrm{emp},\ell,e}(R)$ is fixed and used as the
sensor--environment range-response prior for rendering.

\subsubsection{Empirical range-response construction}
Using synchronized LiDAR--IMU streams, LiDAR--inertial
SLAM~\cite{Fast_lio2} registers the scans, after which a separate
tracker associates persistent, approximately homogeneous planar
surface patches across scans. These recordings track non-checkerboard environmental surfaces and
use no checkerboard labels, while observing the same surface over a
range of distances.

Within each sensor--environment group $(\ell,e)$, tracked-plane
measurements are aggregated into 0.25-m range bins. Subtracting the
median log incidence cosine from the median log intensity gives
$z_{j,b}=a_j+g_b+\epsilon_{j,b}$ for track $j$ and bin $b$, where $a_j$
absorbs surface-specific scale and $g_b$ describes shared range
dependence. We fit $g_b$ with a log-domain cubic P-spline and retain the
central 50\% of stable, range-supported tracks under response-envelope
and adjacent-bin limits. Applying a common population-center scale at
3~m to $\exp(z_{j,b}-\widehat a_j)$ gives $G_{j,\ell,e}(R_b)$.

At each supported range bin, the empirical-mean range-response prior
is computed as the arithmetic mean of the available per-track responses:
\begin{equation}
\bar{G}_{\mathrm{emp},\ell,e}(R_b)
=\frac{1}{|\mathcal{J}_{\ell,e,b}|}
\sum_{j\in\mathcal{J}_{\ell,e,b}}G_{j,\ell,e}(R_b),
\end{equation}
where $\mathcal{J}_{\ell,e,b}$ contains retained tracks supporting $R_b$.
The binned mean is linearly interpolated and mildly smoothed with a
log-domain cubic P-spline to obtain the fixed
$\bar{G}_{\mathrm{emp},\ell,e}(R)$ used for target-reflectivity
estimation and synthetic rendering.

\subsubsection{Target reflectivity priors}
Using the empirical range-response prior, we model the
noise-free intensity of checker class
$c\in\{\mathrm{white},\mathrm{black}\}$ as
\begin{equation}
    I_c(R,\alpha)
    =
    \rho_c \cos\alpha\,
    \bar{G}_{\mathrm{emp},\ell,e}(R),
\end{equation}
where $\rho_c$ is the target-specific reflectivity prior and $\alpha$
is the incidence angle. This is an empirical rendering approximation
for the measured sensor--environment group. For each measured LiDAR return $k$ whose ray intersects the physical
board support in the \emph{good frame}, the normalized reflectivity is
computed as
\begin{equation}
    \widehat{\rho}_k = \frac{I_k}{\cos(\alpha_k)\,\bar{G}_{\mathrm{emp},\ell,e}(R_k)}.
\end{equation}

The white and black reflectivity priors are then estimated robustly as
\begin{equation}
    \rho_c=\operatorname{median}_{k\in\mathcal{S}_c}\widehat{\rho}_k,
\end{equation}
where $\mathcal{S}_c$ contains valid board returns assigned to checker
class $c$, with invalid ranges and near-grazing rays excluded. Thus, $\rho_{\mathrm{white}}$, $\rho_{\mathrm{black}}$, and the checker
alignment are inferred from the single \emph{good frame}.

\begin{figure}[t]
    \centering
    \includegraphics[width=\linewidth, trim={6cm 0 2cm 0}, clip]{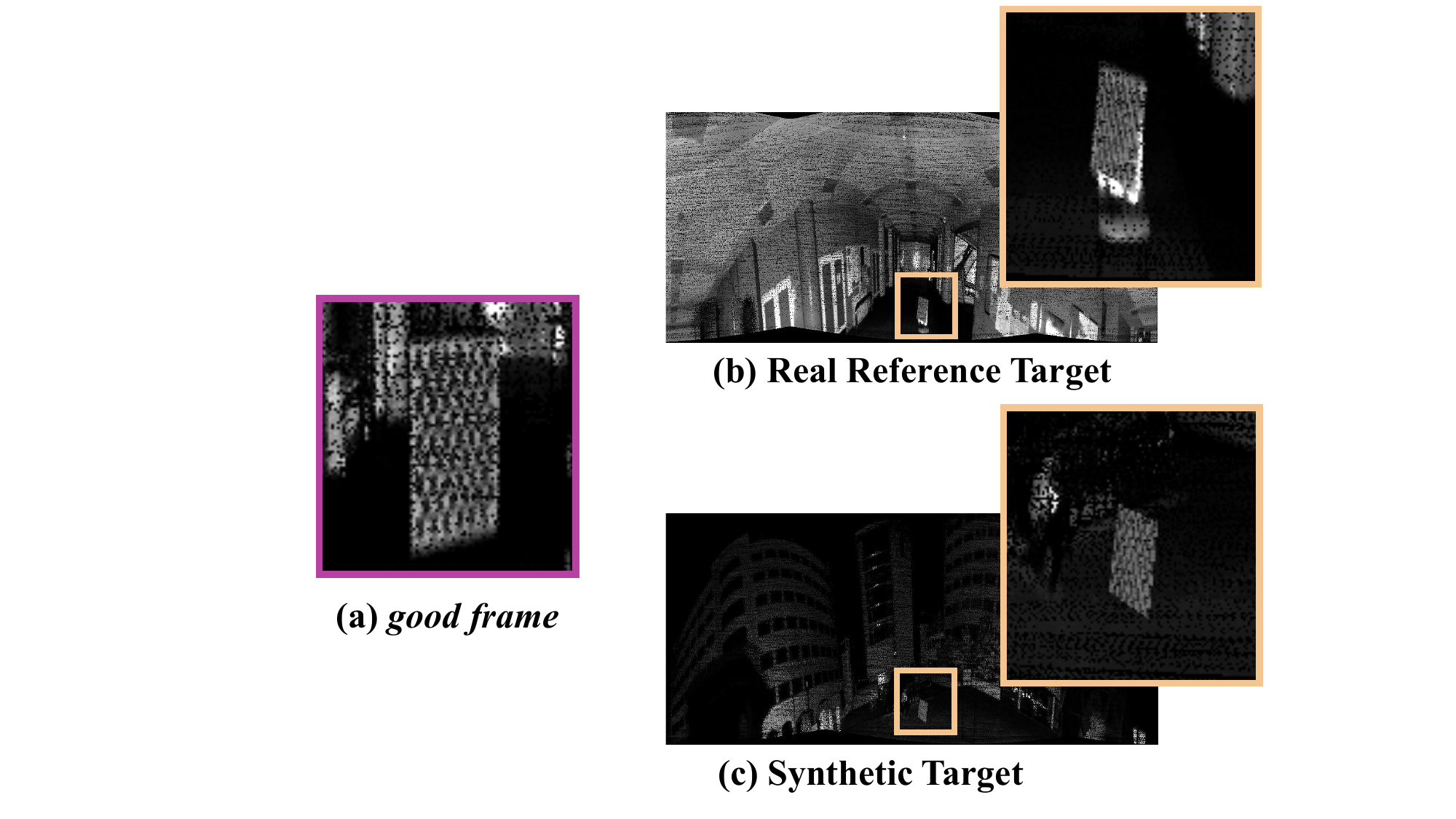}
    \caption{Qualitative comparison of pose-conditioned synthetic rendering.
    (a)~The single annotated observation used to estimate the target-specific
    reflectivity and checker-alignment priors.
    (b)~A real target observed at a different target pose.
    (c)~A synthetic target rendered at the same target pose as (b),
    but composited onto a different background using the priors
    estimated from (a).}
    \label{fig:matched_rendering}
\end{figure}

\subsubsection{Synthetic rendering}
For synthetic observation $m$, a virtual board pose
${}^{L}\mathbf{T}_{B}^{(m)}$ is generated by randomly sampling the board
range, azimuth, and orientation while retaining the physical board
width, height, and checker-square size. Each cylindrical pixel
corresponds to a LiDAR ray. The valid intersection determines the actual range
$R_i^{(m)}$, incidence angle $\alpha_i^{(m)}$, and checker class
$c_i^{(m)}\in\{\mathrm{white},\mathrm{black}\}$. The checker class is
obtained using the fitted in-plane offsets and binary phase.
The noise-free ray-wise synthetic intensity is
\begin{equation}
    I_{\mathrm{syn},i}^{(m)} = \rho_{c_i^{(m)}} \cos\!\left(\alpha_i^{(m)}\right) \bar{G}_{\mathrm{emp},\ell,e} \!\left(R_i^{(m)}\right).
\end{equation}

Although the synthetic board is positioned at a nominal center range,
the actual ray--board intersection distance varies across the board,
especially for oblique orientations. Syn50 targets are composited onto
the Good1 image after removing its original board.

We subsequently add small ray-wise intensity noise and stochastically
remove a subset of the valid board returns to model measurement noise
and return dropout. Thus, the
range-response prior remains deterministic, while the synthetic
observations vary through board pose, ray-wise geometry, intensity
noise, and return dropout. As illustrated in
Fig.~\ref{fig:matched_rendering}, the synthetic target rendered at the
same pose as the real observation closely resembles the real
measurement.

\subsection{One-Round Self-Training}
\label{sec:self_training}

Although ray-aware augmentation and initial training expand the training distribution from a
single annotated observation, synthetic samples cannot fully capture the
variability of real measurements. We therefore use one-round
self-training to incorporate additional reliable real observations
without extra manual annotation~\cite{Lee2013Pseudolabel}.

The detector trained with the initial real and synthetic samples is
applied to an unannotated pool, retaining only the highest-confidence
prediction per observation. A prediction is considered for promotion
only if it falls within the top 30\% and has a confidence score of at
least 0.25. The remaining candidates are further filtered using
physics-consistency checks on range support, projected target size,
valid-return support, and reflectivity contrast.

For each promoted observation, additional ray-aware synthetic
samples are generated and added together with the promoted observation,
yielding the R1 detector after one additional training round. We use a
single self-training round because additional rounds provide limited gains.

\subsection{Detector-Guided 3D Target Localization and Calibration}

Within the detected region of interest, the dominant range cluster is
selected and its plane and normal are estimated using RANSAC. The
bounding-box-center ray intersects the plane to initialize the board
center. A rectangle with the known board dimensions is then fitted to
the detected bounding box to initialize the center and in-plane
orientation. Grid search refines the planar center offsets and rotation
using the Good1 checker phase, white/black reflectivity priors, and
$G(R)$ to maximize reflectivity-pattern consistency. Only poses with
sufficient range support, plane consistency, and physical coverage are
retained as ${}^{L}\widehat{\mathbf{T}}_{B}$.

The corresponding camera-side pose is independently estimated from the
detected checkerboard corners using PnP with the calibrated camera
intrinsics and known board geometry. Only observations with valid poses
in both modalities are used for calibration. Given the LiDAR center and
normal, $\widehat{\mathbf{c}}^{L}_{i}$ and
$\widehat{\mathbf{n}}^{L}_{i}$, and their camera-side counterparts,
$\mathbf{c}^{C}_{i}$ and $\mathbf{n}^{C}_{i}$, we estimate the
camera--LiDAR extrinsic transformation
${}^{C}\mathbf{T}_{L}
=({}^{C}\mathbf{R}_{L},{}^{C}\mathbf{t}_{L})$ such that
\begin{equation}
\mathbf{c}^{C}_{i}
\simeq
{}^{C}\mathbf{R}_{L}\widehat{\mathbf{c}}^{L}_{i}
+{}^{C}\mathbf{t}_{L},
\qquad
\mathbf{n}^{C}_{i}
\simeq
{}^{C}\mathbf{R}_{L}\widehat{\mathbf{n}}^{L}_{i}.
\label{eq:cross_modal_pose_alignment}
\end{equation}
The transformation is obtained by jointly minimizing the center and
normal alignment residuals over all fitting observations.

\section{Experiments}
\label{sec:experiments}

Our sensor rig comprises
a Livox Mid-360, a Velodyne VLP-16, and a Grasshopper camera rigidly
mounted together. The Mid-360 is used for the main checkerboard detection
and camera--LiDAR calibration experiments. The concurrently recorded
VLP-16 data are used to evaluate applicability under a substantially
sparser LiDAR sampling pattern.

The checkerboard measures $W_B = 0.36$~m by $H_B = 0.27$~m, with
$d = 0.03$~m.
Mid-360 scans are projected into cylindrical reflectivity images using a
fixed field of view of
$\theta \in [-180^\circ, 180^\circ]$ and
$\phi \in [-7^\circ, 52^\circ]$, with an image resolution of
$(W_I,H_I)=(1024,512)$. Each main Mid-360 observation aggregates an
approximately 5-s stationary capture, retaining the nearest return per
cylindrical pixel without combining different target poses or sessions.
The final evaluation comprises four independently
captured locations: Indoor1, Indoor2, Outdoor1, and Outdoor2, containing
59, 64, 56, and 63 observations, respectively. Indoor1 and Outdoor1 have relatively open and less cluttered
backgrounds, whereas Indoor2 and Outdoor2 contain more nearby objects
and greater variations in scene structure and reflectivity. Outdoor2 additionally contains dense vegetation, adding further
background complexity.

For each location, one of 20 predefined Good1 candidates is labeled;
the other 19 plus 15/6/6/6 additional unlabeled observations form R1
pools of 34/25/25/25 for Indoor1/Indoor2/Outdoor1/Outdoor2.
Validation sets contain 10/10/5/10 observations. Good1, the R1 pool,
validation, and test are disjoint at the observation level; test
observations, including later additions, are never used for selection
or training.

All detectors are initialized from the same pretrained YOLO11n~\cite{jocher2024yolo11} checkpoint
and trained for 100 epochs. The $1024\times512$ cylindrical images are
downsampled by a factor of two and padded vertically to form
$512\times512$ inputs without geometric distortion.

\subsection{Training conditions and terminology}
\textbf{Good1} uses one labeled frame; \textbf{Good1+Syn50 (R0)} adds 50
ray-aware synthetic samples; \textbf{Good1+Syn50+R1} adds one round of
self-training. Unless specified, \textbf{RAPA} denotes the last
configuration. In the controlled comparison, image-space baselines and
R0 share the same predefined Good1, test split, base training settings,
and 50 repeated or augmented samples per location. A separate
20-candidate analysis measures Good1 sensitivity with fixed splits and
settings.

\subsection{Detection and Augmentation Results}
\label{sec:detection_augmentation}

\begin{table}[t]
    \centering
    \caption{Detection performance under image-space augmentation,
    Good1 initialization sensitivity, and full supervision. Sensitivity
    results are the mean and sample standard deviation over 20 Good1
    candidates; all other results are single runs. Syn50/GT denotes 50
    synthetic samples generated per positive GT observation. Values are
    mAP@0.5/mAP@0.5:0.95.}
    \label{tab:augmentation_comparison}
    \begingroup
    \footnotesize
    \setlength{\tabcolsep}{2.0pt}
    \renewcommand{\arraystretch}{0.8}
    \setlength{\aboverulesep}{0.20ex}
    \setlength{\belowrulesep}{0.20ex}

    \begin{tabular*}{\columnwidth}{@{\extracolsep{\fill}}llcccc@{}}
        \toprule
        \multicolumn{2}{l}{Training condition}
        & Indoor1 & Indoor2 & Outdoor1 & Outdoor2 \\
        \midrule
        \multicolumn{6}{l}{\textit{Predefined Good1 (single run)}} \\
        \multicolumn{2}{l}{Good1 repeated (no aug.)}
        & .405/.119 & .097/.048 & .198/.056 & .022/.010 \\
        \multicolumn{2}{l}{HSV jitter Aug50}
        & .495/.141 & .102/.047 & .178/.066 & .065/.029 \\
        \multicolumn{2}{l}{Horizontal flip Aug50}
        & .538/.174 & .132/.076 & .292/.087 & .059/.027 \\
        \multicolumn{2}{l}{Deep ChArUco Aug50}
        & .623/.285 & .413/.182 & .563/.206 & .167/.056 \\
        \multicolumn{2}{l}{Good1+Syn50 (R0)}
        & .784/.524 & .844/.489 & .779/.565 & .483/.290 \\
        \midrule
        \multicolumn{6}{l}{\textit{Good1 sensitivity (20 candidates)}} \\
        Good1+Syn50 & Mean
        & .752/.434 & .492/.284 & .647/.442 & .206/.120 \\
        & $\pm$ Std.
        & .145/.121 & .331/.195 & .122/.086 & .218/.133 \\
        \rowcolor{black!7}
        Good1+Syn50+R1 & Mean
        & .875/.530 & .652/.383 & .796/.547 & .240/.141 \\
        \rowcolor{black!7}
        & $\pm$ Std.
        & .083/.074 & .253/.164 & .105/.072 & .260/.160 \\
        \midrule
        \multicolumn{6}{l}{\textit{Fully supervised references}} \\
        \multicolumn{2}{l}{All train GT (real only)}
        & .993/.659 & .683/.441 & .757/.484 & .372/.240 \\
        \multicolumn{2}{l}{All train GT + Syn50/GT}
        & .870/.636 & .961/.669 & .772/.604 & .711/.436 \\
        \bottomrule
    \end{tabular*}
    \endgroup
\end{table}

Table~\ref{tab:augmentation_comparison} separates the fixed-Good1
augmentation comparison from initialization sensitivity. HSV jitter and
horizontal flipping are image-space baselines~\cite{Shorten2019}; the
Deep ChArUco-style baseline follows its homography and pixel-degradation
policy~\cite{Hu2019DeepCharuco}. With the same predefined Good1, R0
exceeds Deep ChArUco by 16.1, 43.1, 21.6, and 31.6 percentage points
in mAP@0.5 for Indoor1, Indoor2, Outdoor1, and Outdoor2, respectively.
This comparison evaluates the combined ray-aware augmentation model
before self-training. The separate 20-candidate block reports
initialization sensitivity and the mean change after R1.

Under full supervision, augmentation improves performance in Indoor2
and Outdoor2, although gains are not uniform across locations.
With the predefined Good1 and fixed promotion settings, mean mAP@0.5
over the four locations is .723/.817/.818/.832 for R0/R1/R2/R3,
respectively. Later gains are smaller and non-monotonic across
locations; R1 is therefore used as a fixed choice for the final pipeline.

\subsection{Robustness to Initialization and Operating Conditions}
\label{sec:robustness}

\begin{figure}[t]
    \centering
    \includegraphics[width=\columnwidth]{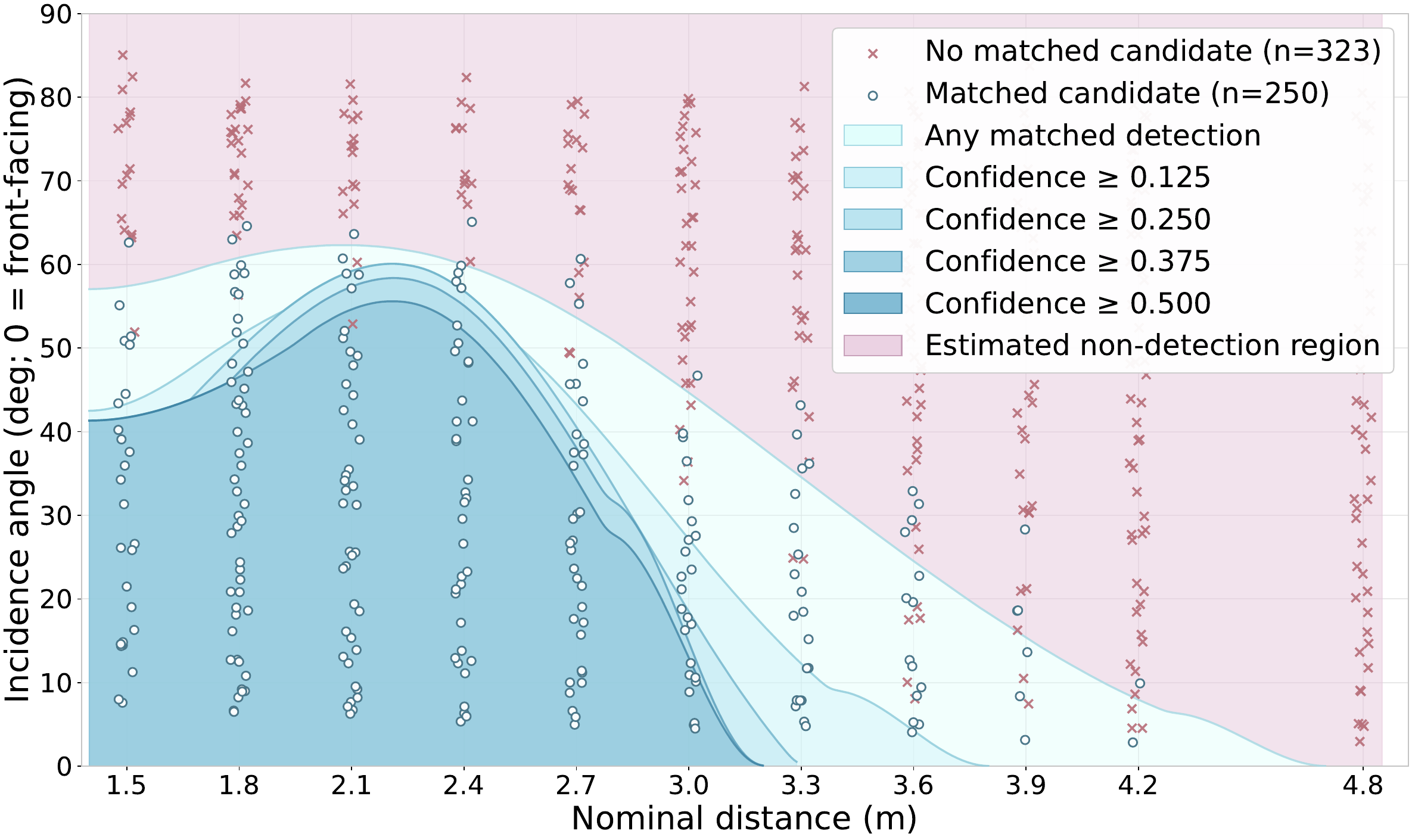}
    \caption{Operating range of the proposed detector with respect to target distance and incidence angle, shown at varying confidence thresholds.}
    \label{fig:range_incidence}
\end{figure}

We evaluate the robustness of RAPA to Good1 selection, location changes,
and target observation geometry.

\subsubsection{Good1 sensitivity}
For each location, 20 randomly sampled Good1 candidates are fixed
before evaluation. Each initializes the same Good1+Syn50+R1 pipeline;
R0 and R1 are evaluated on the same location-specific test set with
the remaining pipeline settings fixed. Table~\ref{tab:augmentation_comparison}
reports the mean and sample standard deviation over all candidates,
without retrospectively selecting the best initialization.
R1 improves mAP@0.5 for 18/20, 19/20, 19/20, and 6/20 candidates in
Indoor1, Indoor2, Outdoor1, and Outdoor2, respectively, and reduces
the standard deviation in the first three locations.
Outdoor2 remains sensitive to initialization: R1 yields
.240$\pm$.260 mAP@0.5, spanning .000--.673 across candidates.
Thus, one annotation can initialize the framework, but its quality
remains important under difficult visibility and background conditions.

\subsubsection{Range and incidence angle}
Figure~\ref{fig:range_incidence} reports detection performance as a
function of target range and incidence angle.
Performance remains stable under well-sampled operating conditions and
degrades as increasing range or oblique incidence reduces the target's
LiDAR support. The detectable region is widest around 2.1\,m, where candidates are
matched even under oblique incidence. Beyond this range, the maximum
detectable incidence angle decreases steadily with distance.

\subsection{Target Detection and Camera--LiDAR Calibration}
We quantitatively compare the complete pipeline against two
training-free target-based calibration methods, Zhou et
al.~\cite{Zhou2018automatic} and ACSC~\cite{cui2020acsc}, measuring
target detection under our small-board, sparse-observation conditions
and calibration obtained from each method's verified valid detections.

\begin{table}[t]
    \centering
    \caption{Comparison with training-free target-based baselines.
     ``Returned'' counts observations in which each method returns a board
    candidate. For RAPA, ``Valid'' requires successful pose recovery and
    center and normal errors of at most 10~cm and $5^\circ$, respectively,
    relative to the manually annotated 3D board pose.
    Each method is calibrated from its
    valid detections, and only when at least three are available.
    $E_r$ and $E_t$ are cross-modal target-normal and target-center
    residuals (median/RMSE), evaluated using the fitted extrinsic
    transformation and including fitting observations.}
    \label{tab:pose_recovery}
    \begingroup
    \footnotesize
    \setlength{\tabcolsep}{2.0pt}
    \renewcommand{\arraystretch}{0.8}
    \setlength{\aboverulesep}{0.20ex}
    \setlength{\belowrulesep}{0.20ex}
    \begin{tabular*}{\columnwidth}{@{\extracolsep{\fill}}llcccc@{}}
        \toprule
        Dataset
        & Method
        & Returned
        & Valid$\uparrow$
        & $E_r$ ($^\circ$)$\downarrow$
        & $E_t$ (cm)$\downarrow$ \\
        \midrule
        Indoor1
        & RAPA
        & 50/59
        & \textbf{46/59}
        & 1.37/5.14
        & 2.55/3.25 \\
        & Zhou et al.
        & 27/59
        & 21/59
        & 1.60/5.10
        & 2.40/2.98 \\
        & ACSC
        & \textbf{51/59}
        & 38/59
        & 1.39/5.12
        & 3.10/3.54 \\
        \midrule
        Indoor2
        & RAPA
        & \textbf{49/64}
        & \textbf{47/64}
        & 1.67/3.20
        & 3.34/3.79 \\
        & Zhou et al.
        & 8/64
        & 1/64
        & --
        & -- \\
        & ACSC
        & 32/64
        & 12/64
        & 1.66/3.21
        & 3.41/3.84 \\
        \midrule
        Outdoor1
        & RAPA
        & 44/56
        & \textbf{44/56}
        & 1.49/2.45
        & 2.31/2.73 \\
        & Zhou et al.
        & 2/56
        & 2/56
        & --
        & -- \\
        & ACSC
        & \textbf{52/56}
        & 39/56
        & 1.33/2.44
        & 3.94/4.62 \\
        \midrule
        Outdoor2
        & RAPA
        & 25/63
        & \textbf{25/63}
        & 1.57/2.27
        & 1.87/1.99 \\
        & Zhou et al.
        & 11/63
        & 2/63
        & --
        & -- \\
        & ACSC
        & \textbf{56/63}
        & 10/63
        & 1.70/2.51
        & 3.15/3.31 \\
        \bottomrule
    \end{tabular*}
    \endgroup
\end{table}
As shown in Table~\ref{tab:pose_recovery}, 162/168 RAPA returns are valid
(96.4\%), yielding target localizations in 162/242 observations (66.9\%);
Zhou et al.\ and ACSC yield 26/242 and 99/242, respectively. 

Extrinsics are fitted from observations with valid LiDAR and camera PnP
poses. $E_r$ and $E_t$ measure target-normal angular and target-center
distance disagreement on these same observations, so they are fitted
cross-modal residuals, not held-out errors. RAPA LiDAR poses come from
the proposed recovery stage, camera poses from corner detection and PnP;
manual 3D annotations determine validity only.

Zhou et al.\ rarely returns a candidate outside Indoor1, with most
failures occurring during corner extraction under non-repetitive LiDAR
sampling. ACSC returns candidates more frequently, but many correspond
to non-target planes satisfying similar geometric constraints; in
Outdoor2, only 10 of 56 returns are valid. Its intensity-pattern
verification also appears less effective at our board scale, where the
3\,cm squares are below the recommended 8\,cm minimum.

For calibration, target-normal residuals are broadly comparable when
calibration is available, whereas RAPA achieves lower target-center
residuals than ACSC in all four locations. Zhou et al.\ provides
insufficient valid detections for calibration outside Indoor1, while
RAPA supports calibration in all four locations.

These results highlight a practical advantage of RAPA: supporting
camera--LiDAR calibration with a compact $0.36 \times 0.27$~m
checkerboard containing $3$~cm squares under sparse LiDAR sampling.
By recovering more valid target observations in this setting, RAPA
improves the availability of usable calibration correspondences.
This capability is particularly relevant to calibration setups
in which target size is constrained.

Target-free MIAS-LCEC~\cite{huang2024mias} returns no estimate on our
single-observation captures, whose projected point count is only
6--10\% of its TF360 data. To also examine conditions compatible with the baselines, we evaluate
the public MIAS-LCEC-CB70 dataset~\cite{huang2024mias}: 15
camera--LiDAR pairs captured using a Mid-70 and a checkerboard with
8\,cm squares at 3--5\,m. ACSC returns 12 candidates, of which 11 are
valid, with median target-normal/center residuals of
$0.59^\circ$/1.22\,cm. MIAS-LCEC returns extrinsic estimates on all
15 pairs, with corresponding medians of $0.47^\circ$/2.54\,cm.
RAPA uses one Good1, seven unlabeled pairs, and seven evaluation pairs,
recovering six valid poses. Since the baselines use all 15 pairs,
these results do not constitute a comparison on identical test sets.
They show that the baselines work under compatible conditions;
our main comparison demonstrates the benefit of reliable LiDAR-side
target localization for calibration under small-board, sparse-observation
conditions, rather than superiority across all sensing conditions.

\subsection{Generalization Across Targets, Sensors, and Locations}
\label{sec:generalization}

To examine generalization beyond the calibration-target configuration
and LiDAR sensor used in the main experiments, we conduct additional
experiments with a larger ChArUco board and a VLP-16. The VLP-16
experiment uses the same concurrently recorded Indoor1 observations as
the Mid-360 experiment, isolating sensor differences from scene
variation. We further examine extension to a circular traffic sign and
cross-location transfer of the checkerboard-derived Good1 prior.
Table~\ref{tab:generalization} summarizes the results.

\begin{table}[t]
    \centering
    \caption{Generalization results using RAPA. ``Valid'' is the count/total
of LiDAR poses satisfying the fixed validity criteria. $E_r$ and $E_t$
are cross-modal target-normal and target-center residuals (median/RMSE).
The final two rows average the four same-location and twelve
cross-location cases within the transfer experiment; Valid is the
unweighted mean of their valid-pose rates.}
    \label{tab:generalization}
    \setlength{\tabcolsep}{2.5pt}
    \resizebox{\columnwidth}{!}{%
    \begin{tabular}{lccccc}
        \toprule
        Experiment
        & mAP@.5$\uparrow$
        & mAP@.5:.95$\uparrow$
        & Valid$\uparrow$
        & $E_r$ ($^\circ$)$\downarrow$
        & $E_t$ (cm)$\downarrow$ \\
        \midrule
        \multicolumn{6}{l}{\textit{Target-configuration transfer}} \\
        Large ChArUco, Outdoor2
        & .682 & .439 & 30/55 & 1.11/1.42 & 3.64/4.36 \\
        Large ChArUco, Indoor2
        & .892 & .621 & 35/55 & 0.97/1.13 & 3.76/4.43 \\
        \midrule
        \multicolumn{6}{l}{\textit{LiDAR-sensor adaptation}} \\
        VLP-16, Indoor1
        & .888 & .391 & 52/59 & 1.85/2.58 & 5.50/6.23 \\
        \midrule
        \multicolumn{6}{l}{\textit{Extension beyond checkerboards}} \\
        Circular traffic sign
        & .993 & .734 & 19/20 & 2.46/5.51 & 1.88/3.35 \\
        \midrule
        \multicolumn{6}{l}{\textit{Cross-location Good1-prior transfer}} \\
        Same-location mean
        & .776 & .500 & 56.8\% & 0.98/2.05 & 2.43/2.64 \\
        Cross-location mean
        & .759 & .468 & 55.5\% & 1.44/2.14 & 2.48/2.91 \\
        \bottomrule
    \end{tabular}%
    }
\end{table}

\subsubsection{Target-configuration transfer}
The additional target is a $0.75 \times 0.55$~m ChArUco board with a
$15 \times 11$ grid and a square size of $0.05$~m. Despite differences
in size, pattern, and material, RAPA recovers valid 3D poses and supports
calibration in both indoor and outdoor environments.

\subsubsection{LiDAR-sensor adaptation}
For VLP-16, we retain the RAPA pipeline with a sensor-specific
range-response prior. Despite its substantially sparser sampling, RAPA
maintains detection performance comparable to Mid-360.

\subsubsection{Extension beyond checkerboards}
We adapt the target geometry and reflectivity priors to a 0.60-m
circular traffic sign. RAPA recovers valid poses for most evaluation
observations and supports calibration. This result shows that the
framework is not restricted to checker patterns and can be extended to
targets with known geometry.

\subsubsection{Cross-location transfer}
This experiment uses separately trained R1 models, rather than the
checkpoints used for the fixed-Good1 results above. Each source-location
detector uses only that location's Good1 and unlabeled pool, then is
frozen for evaluation on all four locations without adaptation or
target-location threshold tuning.
The same-location row in Table~\ref{tab:generalization} averages the
four cases in which source and test locations coincide; the
cross-location row averages the other twelve cases. Within this
comparison, mean mAP@0.5 changes
from .776 to .759 and the mean valid-pose rate from 56.8\% to 55.5\%,
with similar calibration residuals.

\subsubsection{Computational cost}
On an AMD Ryzen 5 7500F and NVIDIA RTX 5060 Ti desktop, online
projection, detection, and pose recovery take 171~ms per observation;
one-time Syn50+R1 adaptation takes 6.7~min. These exclude rosbag/disk
I/O and fixed range-response-prior construction, respectively.

\section{Conclusion}
We presented a ray-aware physics augmentation framework for LiDAR calibration-target detection using only a single labeled target observation. By incorporating ray--board intersection geometry, incidence-angle effects, and an empirical sensor--environment range-response prior, the proposed method generates ray-consistent synthetic target observations across varying poses and distances. Experiments demonstrate improved target detection under the evaluated conditions, with further mean gains from one-round self-training, while the resulting detections support downstream target-pose estimation and camera--LiDAR calibration across all evaluated locations. Additional evaluations across different sensing conditions and structured reflective targets further support the applicability of the proposed framework. Overall, the proposed framework moves target-based camera--LiDAR calibration toward a more automated pipeline by enabling reliable LiDAR-side target localization with minimal manual supervision.

\subsection{Limitations and future work}
The current framework requires one manually annotated target, which limits the degree of automation. Future work could reduce this manual requirement through vision-language-model-assisted target localization and segmentation~\cite{Radford2021CLIP,Liu2024GroundingDINO,Kirillov2023SAM}. Detection reliability also remains sensitive to challenging sensing and background conditions. Incorporating algorithms that exploit temporal information across consecutive observations can improve target detection performance by integrating information over time~\cite{yin2020lidar,koh2023mgtanet}. Additionally, further self-training beyond R1 yields only limited performance gains in our experiments. More advanced pseudo-label promotion strategies may enable more effective use of additional self-training rounds~\cite{liu2021unbiased,yu2022sc}. Finally, while this study focuses on object detection and physics-based augmentation in cylindrical reflectivity images, extending this representation to sequence-level vision tasks such as optical flow~\cite{teed2020raft,vaquero2018hallucinating}, scene flow~\cite{liu2019flownet3d}, and temporal correspondence would be a direction for applying advanced vision algorithms to LiDAR sensing.

\bibliographystyle{IEEEtran}
\bibliography{references}
\end{document}